\documentclass[a4paper,fleqn]{cas-sc}

\hypersetup{
    linkcolor=black,
    urlcolor=black,
    citecolor=black,
    filecolor=black,
    menucolor=black
}

\usepackage{placeins}
\usepackage{subcaption}
\usepackage{tabularx}
\usepackage[numbers]{natbib}
\usepackage{graphicx}
\usepackage{svg}
\usepackage{graphicx}
\usepackage{comment}

\def\tsc#1{\csdef{#1}{\textsc{\lowercase{#1}}\xspace}}
\tsc{WGM}
\tsc{QE}

\begin{document}
\let\WriteBookmarks\relax
\def\floatpagepagefraction{1}
\def\textpagefraction{.001}

\shorttitle{}    

\shortauthors{}  

\title [mode = title]{Token-Based Dual-view Fusion and Adaptation of Large Vision
Models for Breast Cancer
Classification}  

\author[1]{Aysan Ghayouri Pirsoltan}

\ead{aysan_ghayouri@comp.iust.ac.ir}

\credit{}

\affiliation[1]{organization={School of Computer Engineering, Iran University of Science and Technology},
            city={Tehran},
            country={Iran}}

\author[1]{Shima
Babakordi}%[]

\ead{Shima_babakordi@comp.iust.ac.ir}

\credit{}

\author[1]{Mohammad Reza Mohammadi}%[]

\ead{mrmohammadi@iust.ac.ir}

\cormark[1]

\cortext[1]{Corresponding author}

\begin{abstract}
Accurate breast cancer classification from mammography requires effective integration of complementary information from craniocaudal (CC) and mediolateral oblique (MLO) views, which provide a more complete characterization of breast abnormalities. However, existing multi-view learning approaches typically rely on feature-level aggregation or single-stage cross-attention, which can entangle view-specific and shared representations and restrict interaction to limited network depths. To address these limitations, we propose a token-centric dual-view learning framework that unifies prompt-based adaptation and cross-view fusion within a frozen vision transformer backbone. The framework reformulates inter-view interaction as structured token-level communication, where dedicated fusion tokens explicitly encode bidirectional information exchange between CC and MLO views via cross-attention, serving as intermediate carriers of cross-view dependencies rather than relying on direct feature fusion. Unlike conventional methods that apply fusion at a single layer, fusion modules are inserted at multiple transformer depths, enabling progressive and repeated interaction across the encoder hierarchy. Fusion tokens are reintegrated into the token sequence and refined by subsequent transformer layers, facilitating hierarchical propagation of complementary information while preserving view-specific structure. Experiments on VinDr-Mammo and CMMD datasets demonstrate consistent improvements over linear probing, prompt-only adaptation, and conventional fusion baselines. On the VinDr-Mammo BI-RADS classification task, the framework achieves 50.40\% F1-score and 0.8090 AUC, including a 0.10 AUC improvement over a dual-view fusion baseline in the binary setting. Ablation studies further validate the effectiveness of token-based fusion and multi-depth interaction design.
\end{abstract}

\maketitle

\section{Introduction}
Breast cancer remains one of the leading causes of cancer-related mortality worldwide, where early detection through mammography screening plays a critical role in improving survival rates. Despite its clinical importance, automatic interpretation of mammograms remains challenging due to low lesion contrast, complex tissue structures, and subtle inter-class variations. These challenges are further compounded by the need to integrate complementary information across multiple mammographic views. Importantly, clinical diagnosis is inherently multi-view, where radiologists jointly analyze craniocaudal (CC) and mediolateral oblique (MLO) views to exploit complementary anatomical information. However, most existing deep learning approaches do not fully capture this diagnostic process, either relying on single-view analysis or adopting fusion strategies that provide limited cross-view reasoning.

Early deep learning methods primarily focused on single-view mammogram classification using convolutional neural networks, achieving promising results but inherently discarding complementary cross-view information \cite{salama2021deep, cantone2023convolutional, elkorany2023efficient, li2025interpretable}. To address this limitation, multi-view learning strategies have been introduced, typically based on early, intermediate, or late fusion of view-specific representations \cite{lamprou2024stethonet, petrini2022breast, MaMVT}. Early fusion enables information sharing at shallow representation levels, whereas late fusion preserves view-specific feature learning until high-level semantic representations are formed. Intermediate fusion offers a compromise by integrating information during feature extraction. Despite their respective advantages, these strategies generally perform fusion at a single depth of the network, which may limit the ability to capture complementary relationships that emerge across different levels of representation. Consequently, effectively modeling multi-view mammographic information remains an open challenge, as cross-view dependencies are typically captured at isolated depths and are not consistently preserved across hierarchical representations.

More recently, transformer-based architectures have enabled cross-view information exchange through attention-based feature interaction. By allowing representations from one mammographic view to attend to those of another, these methods incorporate complementary anatomical cues during representation learning rather than relying solely on post-hoc feature fusion \cite{brai, MaMVT}. However, in most existing approaches, cross-view attention is implemented as a residual refinement step, where attention outputs are directly added to view-specific representations. This design entangles cross-view and view-specific information, which may limit the explicit modeling and persistence of cross-view cues across layers. As a result, cross-view dependencies are not explicitly preserved as separate representations across deeper layers. In addition, these approaches do not fully leverage the representational capacity of modern vision transformers in a parameter-efficient manner.

Recent medical foundation models have demonstrated strong transferability across diverse tasks, motivating the use of pretrained vision transformers as powerful feature extractors for downstream medical imaging tasks. In parallel, parameter-efficient transfer learning techniques such as visual prompt tuning have emerged as a powerful paradigm for adapting these large pretrained models with minimal trainable parameters \cite{vpt, zhou2022learning}. These methods demonstrate strong performance in low-data regimes while preserving pretrained knowledge. However, existing prompt-based approaches are primarily designed for single-image settings and do not explicitly model structured interactions between multiple correlated medical views.

To overcome these limitations, we propose a two-stage cross-view prompt-driven fusion framework for mammogram classification. First, a deep shared prompt learning strategy is applied across CC and MLO views, enabling parameter-efficient adaptation and cross-view representation alignment while preserving pretrained knowledge. Subsequently, a cross-view fusion mechanism employs bidirectional cross-attention to model complementary information between mammographic views.

Unlike existing cross-view transformer approaches that directly merge attention outputs into feature representations, the proposed framework explicitly preserves cross-view information as dedicated fusion tokens that are appended to the sequence of tokens and refined by subsequent transformer layers. Furthermore, cross-view interaction can be performed at multiple transformer depths, enabling hierarchical integration of complementary information across representation levels. Combined with a final fusion module, the proposed framework unifies early, intermediate, and late multi-view interactions within a single architecture.
\\
\\
\textbf{Contributions}
\begin{itemize}
\item We present a framework for dual-view mammogram classification that integrates representation adaptation and cross-view interaction within a frozen vision transformer backbone.

\item We introduce a deep shared prompt learning strategy that adapts pretrained vision foundation models to multi-view mammography using unified prompt tokens across CC and MLO views, enabling consistent representation alignment while preserving pretrained knowledge.

\item We design a token-based cross-view fusion mechanism that leverages bidirectional cross-attention to encode inter-view dependencies into dedicated fusion tokens, which are reintegrated into the transformer sequence for contextual refinement.

\item We extend cross-view interaction across multiple transformer depths and complement intermediate fusion with a final representation-level fusion, enabling hierarchical multi-level integration of complementary mammographic information.
\end{itemize}

\section{Related Work}
Recent advances in vision-language foundation models have significantly improved representation learning for medical image analysis. Contrastively pretrained models such as BiomedCLIP \cite{zhang2023biomedclip} and MedSigLIP \cite{sellergren2025medgemma} leverage large-scale biomedical image–text pairs to learn transferable visual representations that can be adapted to downstream mammography tasks. Building upon these foundation models, parameter-efficient transfer learning (PETL) methods have emerged as an effective alternative to full model fine-tuning. Approaches including CoOp \cite{zhou2022learning}, Visual Prompt Tuning (VPT) \cite{vpt}, MaPLe \cite{maple}, Tip-Adapter \cite{zhang2021tip}, AdaptFormer \cite{chen2022adaptformer}, and LoRA \cite{hu2022lora} adapt pretrained models by optimizing only a small set of learnable parameters, thereby reducing computational cost while preserving the generalization capability of the pretrained backbone. While these approaches are typically applied to single-image classification tasks, our work extends the prompt-learning paradigm to multi-view mammography classification.

To exploit complementary information across mammographic projections, a wide range of multi-view learning strategies have been proposed. Early approaches mainly relied on image-level, prediction-level, or feature-level fusion. StethoNet \cite{lamprou2024stethonet} investigated multi-view mammography classification using either decision-level fusion through a voting scheme between independently trained CC- and MLO-view networks or image-level fusion by combining both views into a single image before feature extraction. Another approach, DIVF \cite{divf} leverages CC and MLO mammographic views through feature-level fusion within a single backbone network. It combines view-specific feature maps using simple aggregation operations, such as averaging or concatenation, followed by convolutional refinement to obtain a unified representation for classification. CFDV-Net \cite{NGUYEN2026110509} proposes a dual-view mammography framework that processes CC and MLO images through a coarse-to-fine feature extraction pipeline and integrates them using a dedicated Dual Ipsilateral-View Fusion (DIV) module. The fusion alternates between element-wise averaging and feature concatenation followed by convolutional refinement, enabling interaction between complementary view-specific representations for breast cancer classification. While these approaches benefit from information contained in multiple projections, the fusion process typically collapses view-specific information into a single representation, which may limit the model’s ability to preserve explicit relationships between views.

To better handle the multi-instance nature of mammography examinations, weakly supervised and multi-instance learning (MIL) frameworks have also been explored. A representative MIL-based approach for breast cancer prediction in realistic clinical settings \cite{pathak2023weakly} aggregates a variable number of mammographic images per case using different pooling strategies, including mean, max, attention, and gated-attention mechanisms. In addition to instance-level pooling, the same study introduces a domain-specific side-aware aggregation strategy that separately models left and right breast images before performing case-level prediction \cite{pathak2023weakly}. Although MIL-based methods improve flexibility in handling incomplete examinations and variable numbers of views, they rely on permutation-invariant aggregation over instance representations, without explicitly modeling fine-grained interactions between views during representation learning.

Beyond pooling-based fusion strategies, graph-based methods have been proposed to explicitly model cross-view anatomical correspondences in mammography. AGN \cite{agn} constructs region-level graph representations by defining pseudo-landmarks as spatially consistent anatomical anchors across CC and MLO views. These landmarks are used to map spatial features into node representations, forming bipartite graphs that connect corresponding regions between views. Graph convolutional networks are then employed to propagate information across these structured correspondences, enabling reasoning over ipsilateral and contralateral relationships. In addition, an inception-style graph design is introduced to capture multi-scale cross-view dependencies through multiple neighborhood definitions. Although AGN provides a more structured mechanism for modeling inter-view relationships compared to conventional fusion approaches, it relies on handcrafted pseudo-landmark construction and predefined graph connectivity, requiring additional geometric assumptions and preprocessing steps.

More recently, BRAIxMVCCL \cite{brai} introduced a multi-view consistency learning framework that jointly exploits CC and MLO mammographic views through complementary global and local interaction modules. A global consistency module aligns high-level representations across views via cross-view projection and similarity learning, while a local co-occurrence module employs attention-based feature interaction to capture fine-grained cross-view relationships. The resulting global and local representations are concatenated for case-level prediction. Although effective in leveraging complementary information from multiple views, the framework relies on separately designed global and local processing branches with predefined fusion and aggregation mechanisms.

Attention-based architectures and transformer models have further advanced multi-view representation learning by enabling direct interaction between views without requiring explicit registration. 
A cross-view transformer architecture for mammography \cite{van2021multi} introduces cross-attention between intermediate convolutional feature maps of CC and MLO views, enabling bidirectional information exchange between unregistered mammographic projections. 

TransReg \cite{nguyen2023transreg} extends this idea to mammography detection by introducing cross-attention between region proposals from CC and MLO views, enabling region-level interaction and attention-based cross-view alignment across mammographic projections.
MaMVT \cite{MaMVT} integrates cross-view attention within a Swin Transformer \cite{liu2021swin} backbone to enable bidirectional interaction between ipsilateral and contralateral mammographic views at the patch level. TransCorNet \cite{ramakrishnan2025transformer} introduces a transformer-guided cross-view correlation framework for mammography classification that combines lightweight cross-view attention with correlation-based feature alignment to capture both global contextual interactions and fine-grained correspondences between CC and MLO views. Beyond mammography, attention-based fusion has also been explored in other medical imaging applications. A recent attention-driven fusion framework for rotator cuff surgical planning \cite{lee2025multi} performs multi-view fusion between sagittal and coronal MRI using cross-attention between feature embeddings extracted from CNN backbones. Feature maps are projected into token-like embeddings, and cross-attention is used to exchange information between views before aggregation for prediction.

More recently, EAGANet \cite{EAGANet} introduced a dual-stream architecture that jointly processes mammographic projections and breast density information through CNN and state-space feature encoders, and integrates them via attention-based fusion with learnable weight matrix–based state-space modeling. This design enables adaptive interaction between complementary feature streams while effectively capturing both local spatial details and global contextual dependencies. 

Despite the success of attention-driven and transformer-based approaches, most methods integrate cross-view information through residual addition, feature concatenation, or aggregated embeddings, which may lead to entanglement between view-specific and cross-view representations. Furthermore, most existing methods perform cross-view fusion at a single depth or a limited number of layers, whereas our approach enables multi-depth interaction, allowing hierarchical exchange of complementary information across the encoder. Unlike conventional CNN-based or fully fine-tuned transformer pipelines, our framework jointly supports adaptation of a frozen vision transformer and structured cross-view interaction via token-based fusion.

\section{Method}\label{}
\begin{figure*}[pos=htbp]
\centering
\subcaptionbox{Stage 1\label{fig:stage1}}
    {\includegraphics[scale=0.63]{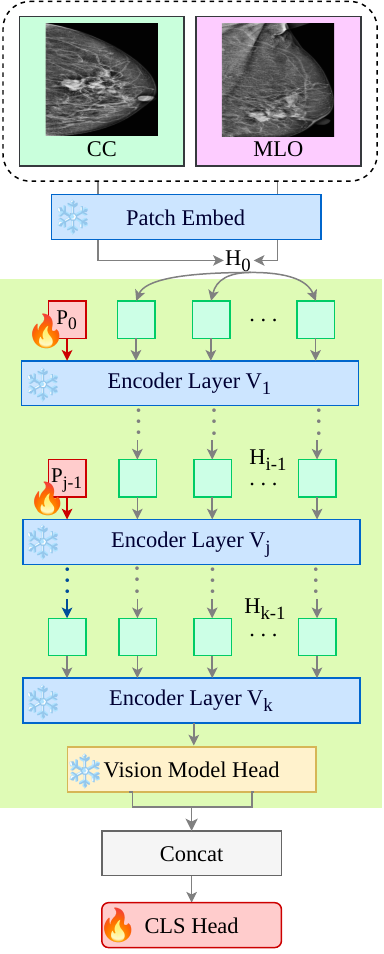}}\hfill
\subcaptionbox{Stage 2\label{fig:stage2}}
    {\includegraphics[scale=0.63]{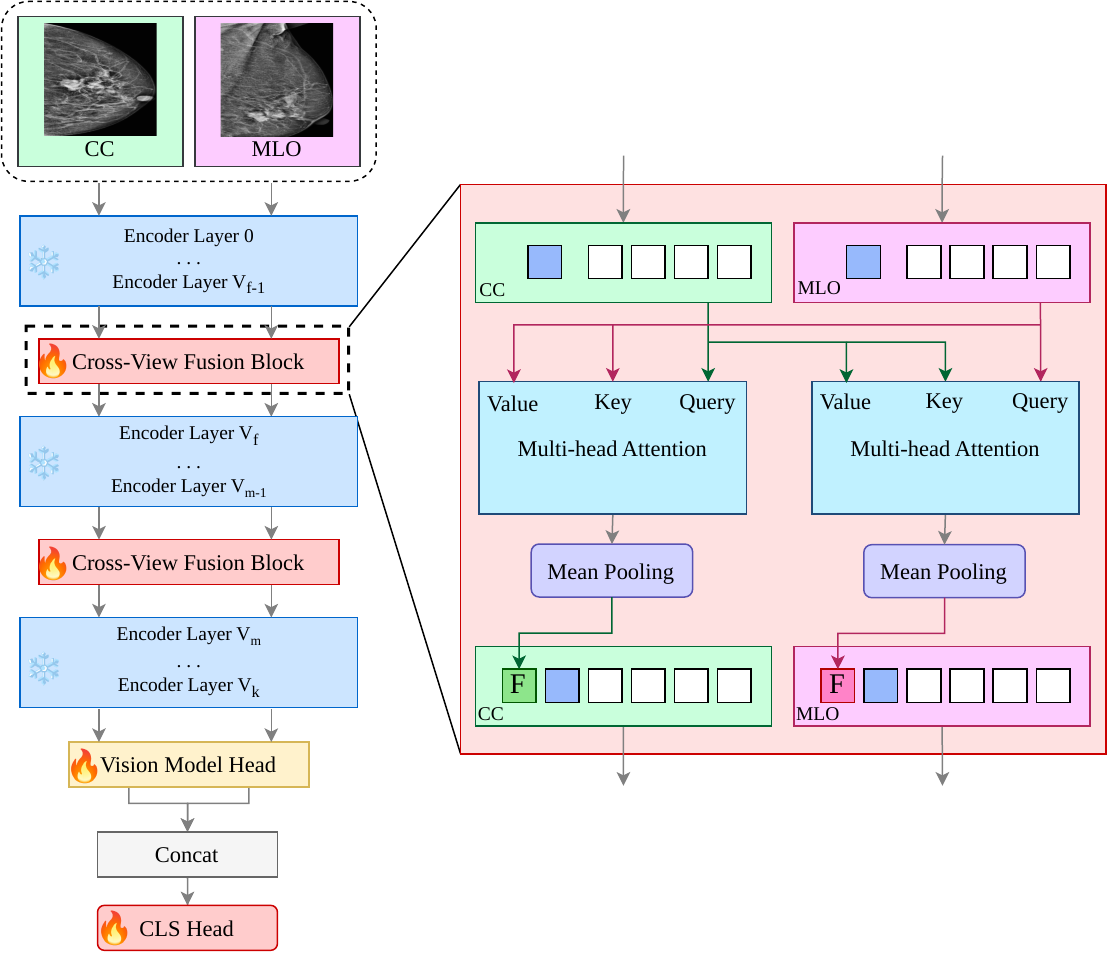}}\hfill

\caption{Overview of the proposed framework. Stage 1 learns view-consistent deep prompts, while Stage 2 introduces cross-view fusion at selected transformer layers. Cross-View Fusion Block enables bidirectional information exchange between CC and MLO views.}
\label{fig:framework}
\end{figure*}

We propose a two-stage framework for dual-view mammogram classification based on a frozen vision foundation encoder MedSigLIP \cite{sellergren2025medgemma}. The first stage performs hierarchical shared-view prompt learning, where learnable prompt tokens are injected across multiple transformer layers to adapt representations from both CC and MLO views within a unified feature space. The second stage introduces cross-view fusion through bidirectional cross-attention, where interaction tokens are inserted into the transformer sequence to enable structured information exchange between views. The final view-level embeddings are concatenated for classification (see Fig. \ref{fig:framework}).

\subsection*{Stage 1: Deep Shared-View Prompt Adaptation}\label{} 
Inspired by Maple \cite{maple} and Visual Prompt Tuning (VPT) \cite{vpt}, we adopt a deep visual prompting strategy in which a small set of learnable prompt tokens is inserted in a pretrained vision Transformer while keeping all backbone parameters frozen. Following the VPT-deep \cite{vpt} formulation, prompt tokens are injected at multiple transformer layers to adapt hierarchical feature representations.

In our setting, this paradigm is extended to dual-view mammography, where craniocaudal (CC) and mediolateral oblique (MLO) images are processed by a shared vision encoder with learnable prompt tokens. A single set of deep visual prompts is jointly optimized and applied to both views, promoting consistent feature adaptation across complementary anatomical perspectives.
\\
Let the pretrained vision transformer consist of $K$ transformer layers. We introduce a set of learnable deep visual prompts
\begin{center}
$\mathcal{P} = \{ \tilde{p}_i \in \mathbb{R}^D \mid i = 1,\dots,J \}$
\end{center}
where $J \leq K$ denotes the prompt depth and $D$ is the hidden dimension \cite{maple}.

A single learnable prompt token is injected at each of the first $J$ transformer layers, while shallow prompting can involve multiple tokens at the input layer, whereas the deep prompts here are exclusively one token per layer.
\\
Given an input image, patch embeddings are first obtained using the frozen embedding layer of the pretrained vision encoder. At each transformer layer $i$, a deep prompt token $\tilde{p}_i$ is added to the input embeddings:
\begin{center}
$H_i = V_i\big( [ \tilde{p}_i, H_{i-1} ] \big),$
\end{center}
where $V_i(\cdot)$ denotes the $i$-th transformer layer and $H_{i-1}$ represents the patch embeddings propagated from the previous layer.

For layers beyond the prompt depth $J$, no prompt tokens are injected and the hidden states are forwarded normally through the frozen transformer layers:
\begin{center}
$H_i = V_i(H_{i-1}), \quad i = J+1, \ldots, K.$
\end{center}

After the final transformer layer, the hidden states are passed through the pretrained multi-head attention pooling head of the vision encoder, producing a compact image-level embedding:
\begin{center}
$z = \mathrm{Head}(H_K), \quad z \in \mathbb{R}^D.$    
\end{center}
Each sample includes CC and MLO views, encoded through a single shared vision encoder with learnable prompt tokens:
\begin{center}
$z_{cc} = f_{\text{vision}}(CC), \quad
z_{mlo} = f_{\text{vision}}(MLO).$
\end{center}
The resulting representations are concatenated to form a joint feature vector:
\begin{center}
$z = [z_{cc}; z_{mlo}] \in \mathbb{R}^{2D}.$
\end{center}
A lightweight linear classifier is trained on top of the concatenated feature representation to predict the final label:
\begin{center}
$\hat{y} = \mathrm{FC}(z).$
\end{center}
During training, only the prompt tokens and classifier parameters are updated, while the backbone vision encoder and the pretrained pooling head remain frozen; the model is optimized using the standard cross-entropy loss.

\subsection*{Stage 2: Token-Based Cross-View Fusion}\label{}

Stage 1 establishes a shared representation space by learning a set of deep prompt tokens jointly optimized across CC and MLO mammograms. By injecting shared prompts into the early layers of the pretrained vision transformer, the model learns a view-consistent adaptation of the backbone while preserving its pretrained knowledge. Nevertheless, the interaction between views remains indirect, as information is exchanged only through the shared prompt parameters.

To further enhance feature adaptation and cross-view integration, we introduce a second training stage based on cross-view fusion tokens (see Fig. \ref{fig:stage2}). Inspired by the token-insertion paradigm of deep prompt learning, the proposed approach extends adaptation beyond the prompt-learning stage by explicitly exchanging information between mammographic views at intermediate levels of the transformer. Rather than directly combining feature maps through summation or averaging, complementary information from one view is summarized into a compact fusion token and injected into the token sequence of the other view. This strategy allows subsequent transformer layers to further refine the representation using cross-view information while maintaining the original transformer architecture.

Let $H_{cc}$ and $H_{mlo}$ denote the intermediate hidden representations of the CC and MLO views, respectively. At selected fusion layers, the two representations interact through a Cross-View Fusion Block (CVFB), which performs bidirectional cross-attention followed by fusion-token generation and insertion. Specifically, each view attends to the feature tokens of the complementary view according to

\begin{center}
$A_{cc} = \mathrm{MHA}(H_{cc}, H_{mlo}, H_{mlo}),$
\end{center}

\begin{center}
$A_{mlo} = \mathrm{MHA}(H_{mlo}, H_{cc}, H_{cc}),$
\end{center}

where $\mathrm{MHA}(\cdot)$ denotes multi-head attention.

The cross-attention outputs are subsequently aggregated into compact fusion tokens via mean pooling over the token dimension, summarizing complementary anatomical information from the opposite view. The resulting fusion tokens are then inserted into the corresponding token sequences, enabling explicit cross-view information propagation within the transformer.
Formally, the fusion tokens are defined as:
\begin{center}
$t_{cc} = \frac{1}{N}\sum_{i=1}^{N} A_{cc}^{(i)}, \qquad
t_{mlo} = \frac{1}{N}\sum_{i=1}^{N} A_{mlo}^{(i)}.$
\end{center}

where \(N\) denotes the number of tokens in the cross-attention output.
The updated token sequences are propagated to subsequent transformer layers:
\begin{center}
$H'_{cc} = [t_{cc}, H_{cc}], \qquad
H'_{mlo} = [t_{mlo}, H_{mlo}].$
\end{center}

This formulation follows the prompt insertion paradigm, allowing the fusion tokens to be jointly processed with the original visual tokens and progressively refined through subsequent transformer layers. Fusion modules are inserted at designated transformer layers, enabling hierarchical cross-view interactions throughout the encoder.

After the final transformer layer, CC and MLO representations are aggregated and concatenated, as in Stage 1, and passed to a linear classifier for prediction. During Stage 2, only the fusion modules, attention pooling head, and classifier are updated, while the pretrained vision backbone and all prompt parameters learned in Stage 1 remain frozen.

\section{Datasets}\label{}
We conducted experiments on two publicly available mammography datasets: VinDr-Mammo~\cite{nguyen2023vindr} and the Chinese Mammography Database (CMMD)~\cite{cai2023online}.

\textbf{VinDr-Mammo:}
VinDr-Mammo consists of 5,000 four-view examinations (20,000 images) with breast-level BI-RADS \cite{spak2017bi} annotations. We used this dataset for both (1) five-class BI-RADS and (2) binary classification. The dataset exhibits a noticeable class imbalance across BI-RADS categories, as summarized in Table~\ref{tab:dataset_split}. For the multi-class task, we adopted the official train/test partition and randomly selected 400 cases from the training set to form the validation set.

For the binary setting, following~\cite{divf}, BI-RADS~2 was treated as \textit{Suspicious Benign}, while BI-RADS~4 and BI-RADS~5 were grouped as \textit{Suspicious Malignancy}. The original test partition was preserved, while approximately 10\% of the training samples were held out to form a validation set.

\textbf{CMMD:}
CMMD contains 5,202 mammography images collected from 1,775 patients with pathology-confirmed diagnoses. Since no official split is provided, the binary classification dataset was divided into training, validation, and test sets using stratified sampling. The resulting benign and malignant sample distributions across the training, validation, and test sets are reported in Table~\ref{tab:dataset_split}.

All dataset partitions were performed at the \textit{patient level} to prevent information leakage between the training, validation, and test sets.
\\
\\
As a preprocessing step, each mammogram was first cropped to the breast region using foreground contour detection in order to remove most of the irrelevant background area. Contrast enhancement was then applied using Contrast Limited Adaptive Histogram Equalization (CLAHE) to improve the visibility of tissue structures \cite{mammobench}. Finally, all images were resized to $448 \times 448$
 pixels to match the MedSigLIP \cite{sellergren2025medgemma} input resolution.

\begin{table}[b]
\centering
\caption{Distribution of samples across the training, validation, and test splits for the VinDr-Mammo and CMMD datasets. VinDr-Mammo is reported using BI-RADS categories, while CMMD is reported using benign and malignant labels.}
\label{tab:dataset_split}

\begin{tabular}{lccccccc}
\toprule
\multirow{2}{*}{\textbf{Class}} 
& \multicolumn{5}{c}{\textbf{VinDr-Mammo}} 
& \multicolumn{2}{c}{\textbf{CMMD}} \\
\cmidrule(lr){2-6} \cmidrule(lr){7-8}
& \textbf{BI-RADS 1} & \textbf{BI-RADS 2} 
& \textbf{BI-RADS 3} & \textbf{BI-RADS 4} & \textbf{BI-RADS 5} 
& \textbf{Benign} & \textbf{Malignant} \\
\midrule
\textbf{Train}       & 4,842 & 1,688 & 327 & 265 & 78 & 472 & 1,697 \\
\textbf{Validation} & 520 & 183 & 45 & 40 & 12 & 18  & 75 \\
\textbf{Test}       & 1341 & 467 & 93 & 76 & 23 & 68  & 271 \\
\bottomrule
\end{tabular}
\end{table}

\section{Experiments}\label{}
\subsection{Training Details}\label{}
All experiments were conducted using the MedSigLIP vision encoder \cite{sellergren2025medgemma} initialized from publicly available pretrained weights on a single NVIDIA RTX 4090 GPU. Class imbalance was handled using an imbalanced sampling strategy during training.

The proposed framework was trained in a two-stage manner. In Stage 1, the vision encoder was frozen, and only the deep prompt parameters and classification head were optimized. Training was performed for 15 epochs using the AdamW \cite{kingma2014adam} optimizer with a learning rate of 1e-3 and weight decay of 0.01. Optimization used cross-entropy loss with a cosine learning-rate schedule and linear warm-up over the first 100 steps, with a batch size of 8.

In Stage 2, the deep prompts learned in Stage 1 were frozen and reused for initialization. Cross-view fusion modules were inserted within the transformer. Each module employed multi-head cross-attention with four heads. Only the fusion modules, vision encoder head, and the classification layer were trained, while the backbone remained frozen. Training followed the same optimization procedure as in Stage 1.

The number and placement of fusion modules and prompt depth were tuned separately for each dataset using the validation set.
\subsection{Evaluation Metrics}\label{}

Model performance was evaluated using the Area Under the Receiver Operating Characteristic Curve (AUC) and the macro-averaged F1-score. Considering the class imbalance in the datasets, the macro F1-score was employed to ensure that the performance of each class contributes equally to the final evaluation, preventing dominance by majority classes. AUC was additionally used to measure the model's discriminative ability across all possible decision thresholds. Together, these metrics provide a comprehensive assessment of model performance under class imbalance. All reported results correspond to the held-out test set.

\subsection{Baseline Methods}\label{}

\textbf{Linear Probing (LP).}
As a strong parameter-efficient baseline, we evaluate a linear probing strategy built upon the pretrained MedSigLIP vision encoder. During training, all parameters of the vision transformer are frozen and only a single linear classification layer is optimized. Each mammographic image is treated as an independent training sample, regardless of its acquisition view (CC or MLO), and the classifier is trained using image-level supervision.

During inference, breast-level predictions are obtained by processing the corresponding CC and MLO mammograms independently through the frozen encoder and the trained linear classifier. Let $\mathbf{p}_{cc}$ and $\mathbf{p}_{mlo}$ denote the class probability vectors predicted from the CC and MLO views, respectively. Two aggregation strategies are considered:

\begin{itemize}
\item \textbf{Average Fusion (LP-Avg):}
\begin{equation*}
\mathbf{p}_{avg}=\frac{\mathbf{p}_{cc}+\mathbf{p}_{mlo}}{2},
\end{equation*}
where the final prediction is obtained from the averaged probability distribution.
\item \textbf{Maximum Fusion (LP-Max):}
\begin{equation*}
\mathbf{p}_{max}=\max(\mathbf{p}_{cc},\mathbf{p}_{mlo}),
\end{equation*}
where the maximum probability for each class is selected across the two views before prediction.

\end{itemize}

This baseline evaluates the representational quality of the frozen MedSigLIP features and provides a reference for measuring the benefits of prompt-based adaptation and explicit cross-view fusion. Unlike the proposed framework, LP does not perform feature interaction between CC and MLO views and relies solely on prediction-level aggregation.

\textbf{Deep Shared-View Prompt Learning (Stage 1).} The model is adapted using the proposed shared-view prompt learning strategy without any explicit cross-view fusion modules.

\textbf{Cross-View Fusion Only.}
To assess the contribution of the proposed shared-view prompt learning stage, we construct a baseline that employs the Stage 2 cross-view fusion modules without initializing the model with the learned prompts from Stage 1. In this setting, intermediate CC and MLO representations interact through the same cross-view attention mechanism used in the proposed framework, but no shared prompt-based adaptation is performed.

\textbf{Full Framework.} The full framework combining Stage 1 shared-view prompt learning and Stage 2 cross-view fusion with trainable interaction tokens.

All baseline methods use the same pretrained backbone, training protocol, data preprocessing pipeline, and evaluation metrics to ensure a fair comparison.

\subsection{Results and Comparison}
\subsubsection{Results on Baselines}

Table \ref{tab:results} reports the performance of the proposed framework baseline configurations on the CMMD and VinDr-Mammo (5-class) datasets.

Compared with linear probing, all adaptation-based methods yield consistent improvements across datasets. On CMMD, the proposed method improves F1-score from 57.57\% to 64.96\% and AUC from 0.7045 to 0.7161. On VinDr-Mammo, it achieves an F1-score of 50.40\% and an AUC of 0.8090, outperforming all baselines by a large margin.

The results further demonstrate the complementary effect of the two training stages. Shared-view prompt learning consistently improves over linear probing, indicating effective domain adaptation of pretrained representations. Cross-view fusion without prompt initialization also yields competitive performance, highlighting the benefit of explicit CC–MLO interaction even in the absence of prompt adaptation.

\begin{figure*}[pos=htbp]
\centering

\begin{subfigure}[t]{0.48\textwidth}
\centering
\includegraphics[width=\textwidth]{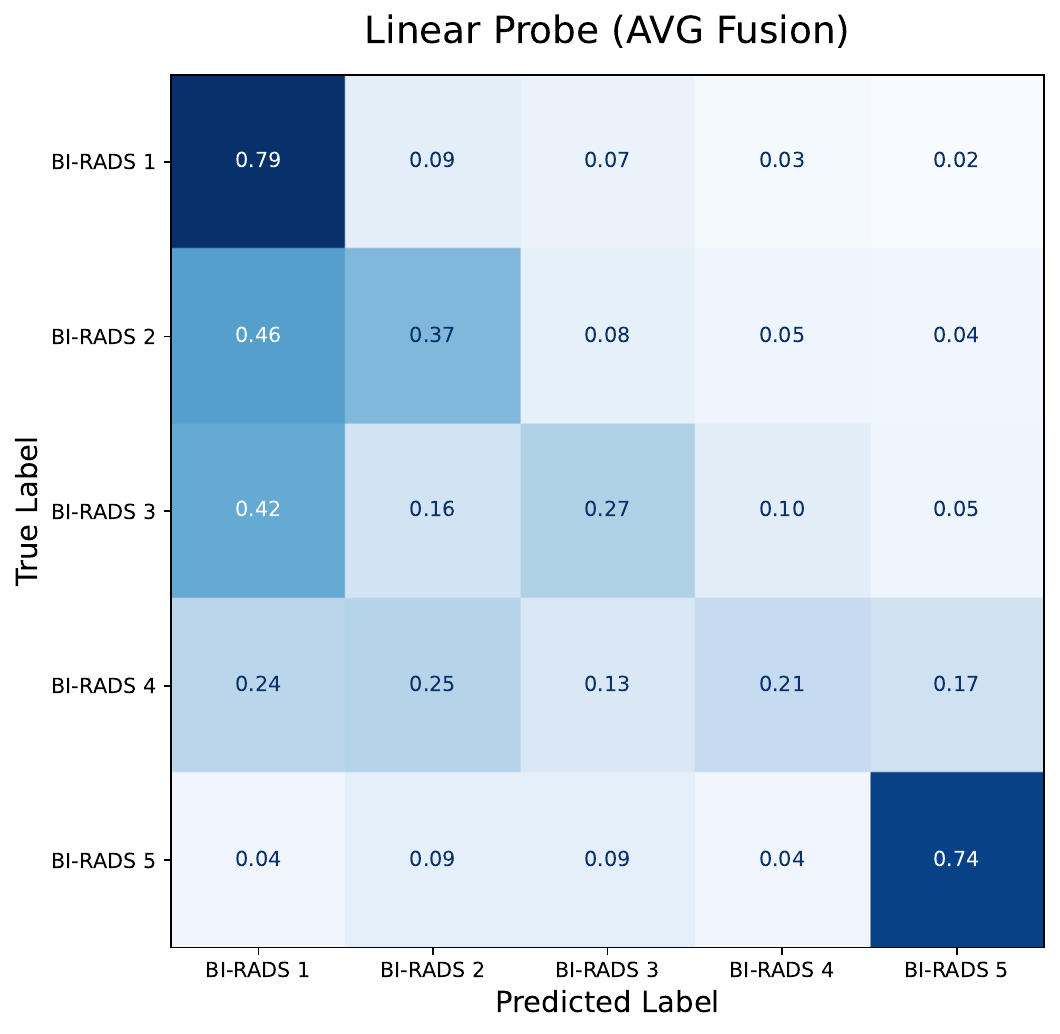}

\label{fig:cm_linear}
\end{subfigure}
\hfill
\begin{subfigure}[t]{0.48\textwidth}
\centering
\includegraphics[width=\textwidth]{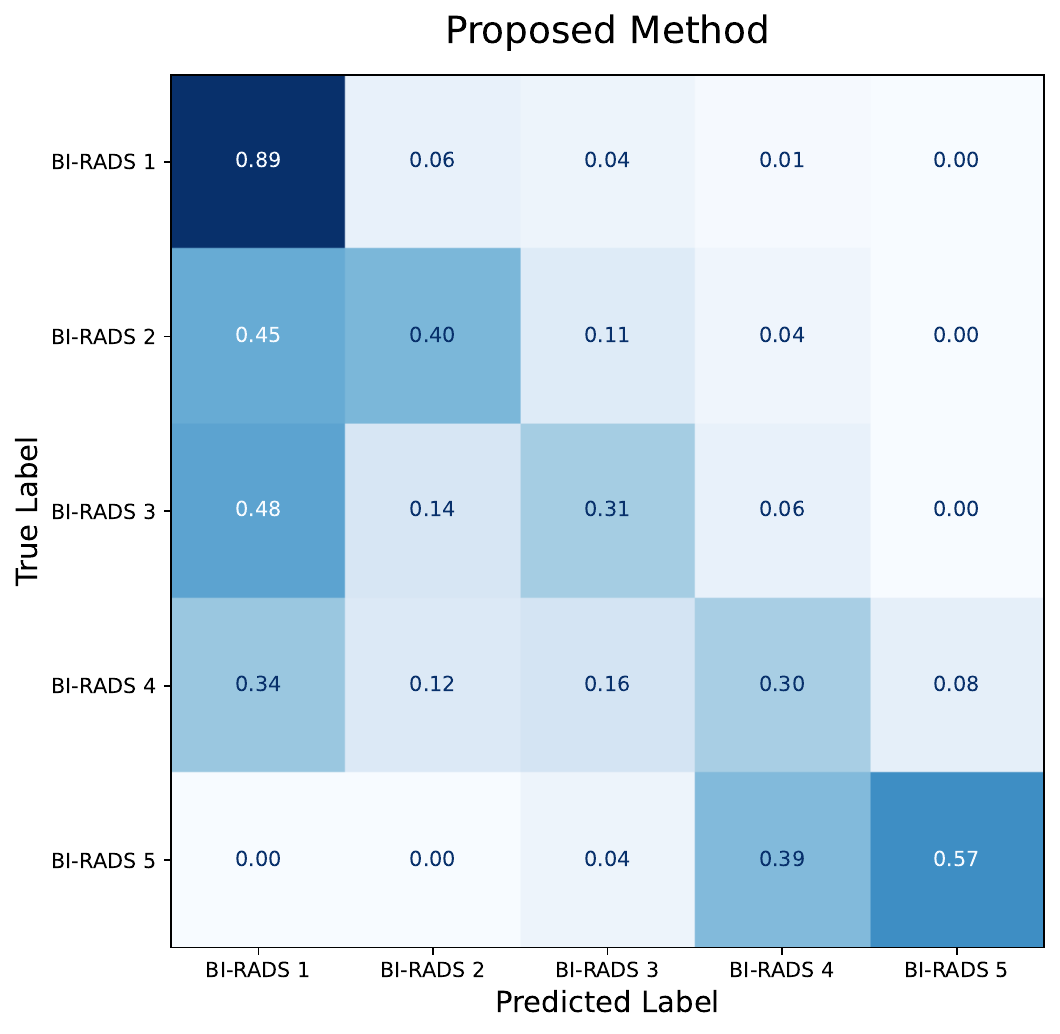}

\label{fig:cm_proposed}
\end{subfigure}

\caption{Normalized confusion matrices on the VinDr-Mammo 5-class test set. Each row is normalized to sum to 100\%. Compared with the linear probe baseline using average fusion (MAE = 0.5575), the proposed method achieves a substantially lower mean absolute error (MAE = 0.3820) and exhibits stronger diagonal dominance, indicating improved classification accuracy and fewer ordinal prediction errors across BI-RADS categories.}
\label{fig:confusion_matrices}
\end{figure*}

Overall, the full framework achieves the most consistent performance across datasets. While the fusion-only variant attains the highest F1-score on CMMD, the proposed model yields the best overall trade-off, achieving the highest AUC on both datasets and the best F1-score on VinDr-Mammo. These results indicate that prompt-based adaptation and cross-view fusion provide complementary benefits for mammography classification. Figure \ref{fig:confusion_matrices} further illustrates the classification behavior on the VinDr-Mammo 5-class test set, demonstrating the impact of the proposed framework on class-wise prediction performance.

\begin{table}[pos=htbp]
\centering
\caption{Performance comparison of different adaptation and fusion strategies on the CMMD and VinDr-Mammo datasets. The best result for each metric is highlighted in bold.}
\label{tab:results}
\begin{tabular}{lcccc}
\hline
& \multicolumn{2}{c}{CMMD} & \multicolumn{2}{c}{VinDr-Mammo (5-Class)} \\
\cline{2-3} \cline{4-5}
Method & F1-Score (\%) & AUC & F1-Score (\%) & AUC \\
\hline
Linear Probe (Avg Fusion) & 57.57 & 0.7045 & 38.95 & 0.7548 \\
Linear Probe (Max Fusion) & 57.57 & 0.7036 & 36.47 & 0.7423 \\
\hline
Stage1: Shared-View Prompt Learning & 61.92 & 0.6984 & 47.66 & 0.7887 \\
Stage2: Cross-View Fusion Only & \textbf{65.20} & 0.7112 & 47.85 & 0.7692 \\
Full Framework  (Stage1 + Stage2) & 64.96 & \textbf{0.7161} & \textbf{50.40} & \textbf{0.8090} \\
\hline
\end{tabular}
\end{table}

\subsubsection{Comparison with Existing Method}

We compare our method with DIVF \cite{divf}, a dual-view mammography framework designed to exploit CC and MLO images for breast cancer classification. DIVF serves as a strong baseline as it explicitly models inter-view relationships using feature-level fusion of the two mammographic views.

Table \ref{tab:divf} presents the performance comparison on the VinDr-Mammo binary classification task. The proposed method achieves consistently higher performance than DIVF \cite{divf} across evaluation settings in terms of F1-score and AUC-ROC.

These results indicate that the proposed cross-view fusion framework provides effective interaction between CC and MLO representations. This is further supported by the use of a hierarchical fusion design with multiple cross-view fusion blocks, enabling interaction at different levels of the encoder rather than a single fusion step.

\begin{table}[cols=4,pos=ht]
\centering
\caption{Comparison with DIVF \cite{divf} under different numbers of cross-view fusion blocks on the VinDr-Mammo binary classification task.}
\label{tab:divf}
\begin{tabular}{lccc}
\hline
Method & \# Fusion Blocks & F1-Score (\%) & AUC-ROC \\
\hline
DIVF \cite{divf} & - & 75.98 & 0.7486 \\
\hline
& 1 & 78.86 & 0.8441 \\
\textbf{Proposed method} & 2 & 77.88 & \textbf{0.8593} \\
& 4 & \textbf{79.57} & 0.8313 \\
\hline
\end{tabular}
\end{table}

\subsection{Ablation Study}
We conduct ablation studies on the CMMD and VinDr-Mammo datasets (VinDr-2-class and VinDr-5-class settings) to evaluate the contribution of each component in the proposed framework. All experiments are performed under identical training settings and are evaluated using F1-score and AUC-ROC.

\textbf{Fusion Strategy.}
We first assess the effect of the proposed fusion mechanism by comparing it with a residual cross-attention fusion baseline. In the baseline, cross-view attention is computed between CC and MLO feature representations, and the resulting features are added to the original token sequence via residual connections. In the proposed method, dedicated fusion tokens are introduced to represent cross-view interactions and are processed through subsequent transformer layers.

As reported in Table~\ref{tab:fusion_strategy}, the proposed method consistently outperforms the residual fusion baseline on both datasets. On VinDr-5, fusion tokens lead to marked performance gains, with increases of 5.22 percentage points in F1-score and 0.0521 in AUC. More modest yet consistent gains are also observed on CMMD.

These results indicate that representing cross-view information via dedicated tokens is more effective than direct residual cross-attention fusion under the same training configuration.

\begin{table}[h]
\centering
\caption{Comparison of fusion strategies. Residual fusion directly adds cross-view attention outputs to the feature sequence, while the proposed method introduces learnable fusion tokens that are processed by subsequent transformer layers.}
\label{tab:fusion_strategy}
\begin{tabular}{lcccc}
\toprule
& \multicolumn{2}{c}{CMMD} & \multicolumn{2}{c}{VinDr-Mammo (5-Class)} \\
\cmidrule(lr){2-3}
\cmidrule(lr){4-5}
Fusion Strategy & F1-Score (\%) & AUC & F1-Score (\%) & AUC \\
\midrule
Additive Residual Fusion & 60.97 & 0.7105 & 45.18 &  0.7569 \\
Proposed Fusion Tokens & \textbf{64.96} & \textbf{0.7161} &
\textbf{50.40} & \textbf{0.8090} \\
\bottomrule
\end{tabular}
\end{table}

\textbf{Fusion Token Aggregation.}
We evaluate different aggregation strategies for fusion tokens, including max pooling, mean pooling, and attention-based pooling. As shown in Table~\ref{tab:fusion_method}, mean pooling achieves the best performance on the VinDr-Mammo Binary setting. Max pooling and attention-based pooling yield lower performance. This suggests that mean pooling provides a more stable aggregation of fusion token representations under the evaluated setting.
\begin{table}[h]
\centering
\caption{Effect of fusion token aggregation strategies on the VinDr-Mammo binary classification task.}
\label{tab:fusion_method}
\begin{tabular}{lcc}
\toprule
Aggregation Method & F1-Score (\%) & AUC-ROC \\
\midrule
Max pooling & 72.13 & 0.8227 \\
Attention pooling & 69.87 & 0.8475 \\
Mean pooling & \textbf{77.88} & \textbf{0.8593} \\
\bottomrule
\end{tabular}
\end{table}

\textbf{Vision Encoder Head Fine-Tuning.}
We study the impact of fine-tuning the vision encoder head. As shown in Table~\ref{tab:vision_head_ft}, fine-tuning improves performance compared to a frozen configuration, with the best results obtained at a learning rate of 1e-5, indicating that partial adaptation of visual features is beneficial.

We refer to the vision encoder head as the projection/pooling module following the transformer backbone. Importantly, even when this module is kept frozen, the proposed method still achieves competitive performance compared to baseline methods. This indicates that the effectiveness of the framework is not solely dependent on fine-tuning of the vision encoder head, and that the proposed fusion and prompt learning components contribute substantially to the overall performance.

\begin{table}[h]
\centering
\caption{Ablation study on vision encoder head fine-tuning for the VinDr-Mammo binary classification task.}
\label{tab:vision_head_ft}
\begin{tabular}{lcc}
\toprule
Training Strategy & F1-Score (\%) & AUC-ROC \\
\midrule
Frozen encoder head & 76.89 & 0.8338 \\
Learning rate = 1e-7 & 76.12 & 0.8468 \\
Learning rate = 1e-5 & \textbf{77.88} & \textbf{0.8593} \\
\bottomrule
\end{tabular}
\end{table}

\begin{figure}[pos=htbp]
\centering

\begin{subfigure}[t]{0.48\linewidth}
    \centering
    \includegraphics[height=6.5cm,keepaspectratio]{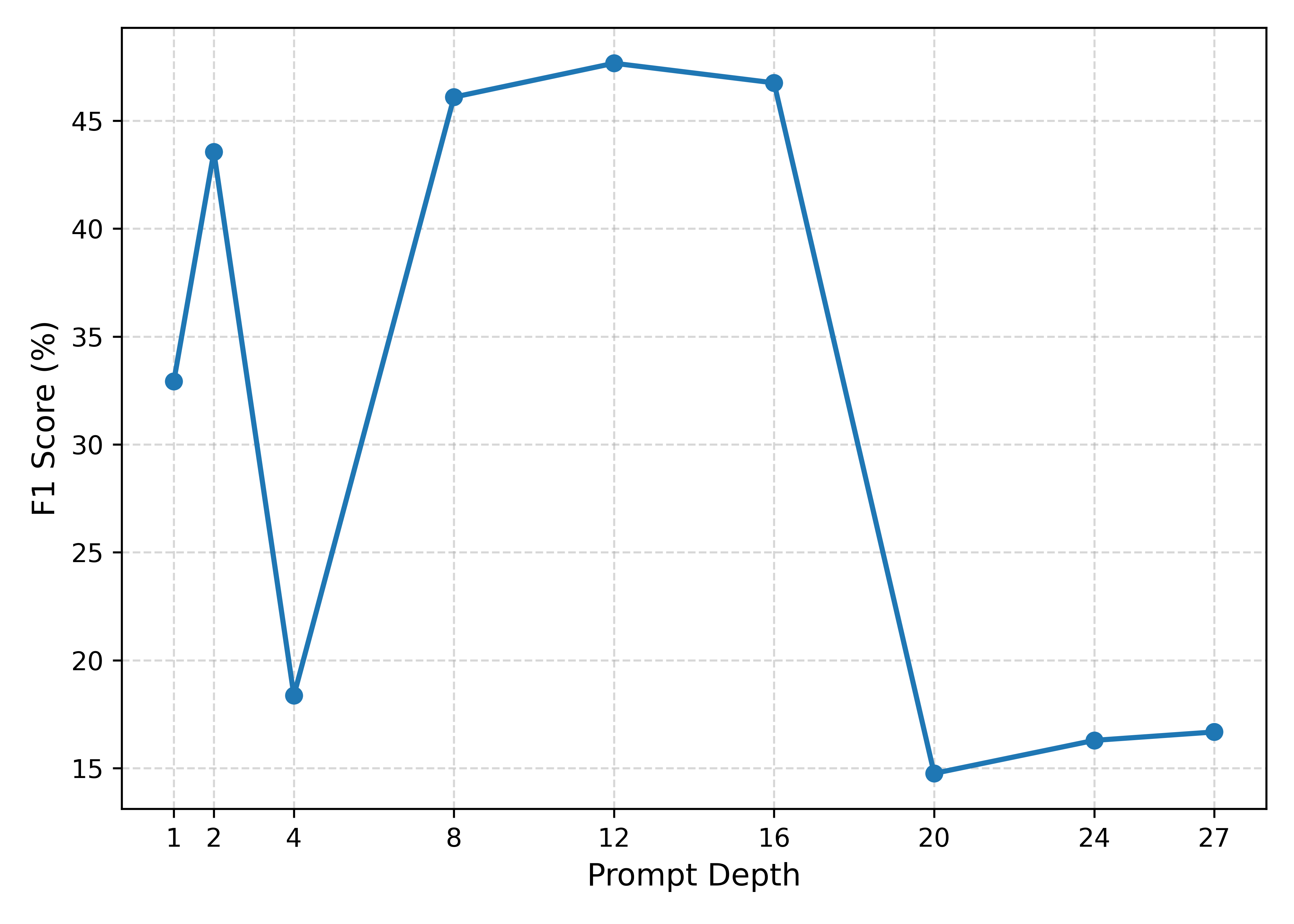}
    \caption{}
    \label{fig:prompt_depth}
\end{subfigure}
\hfill
\begin{subfigure}[t]{0.48\linewidth}
    \centering
    \includegraphics[height=6.5cm,keepaspectratio]{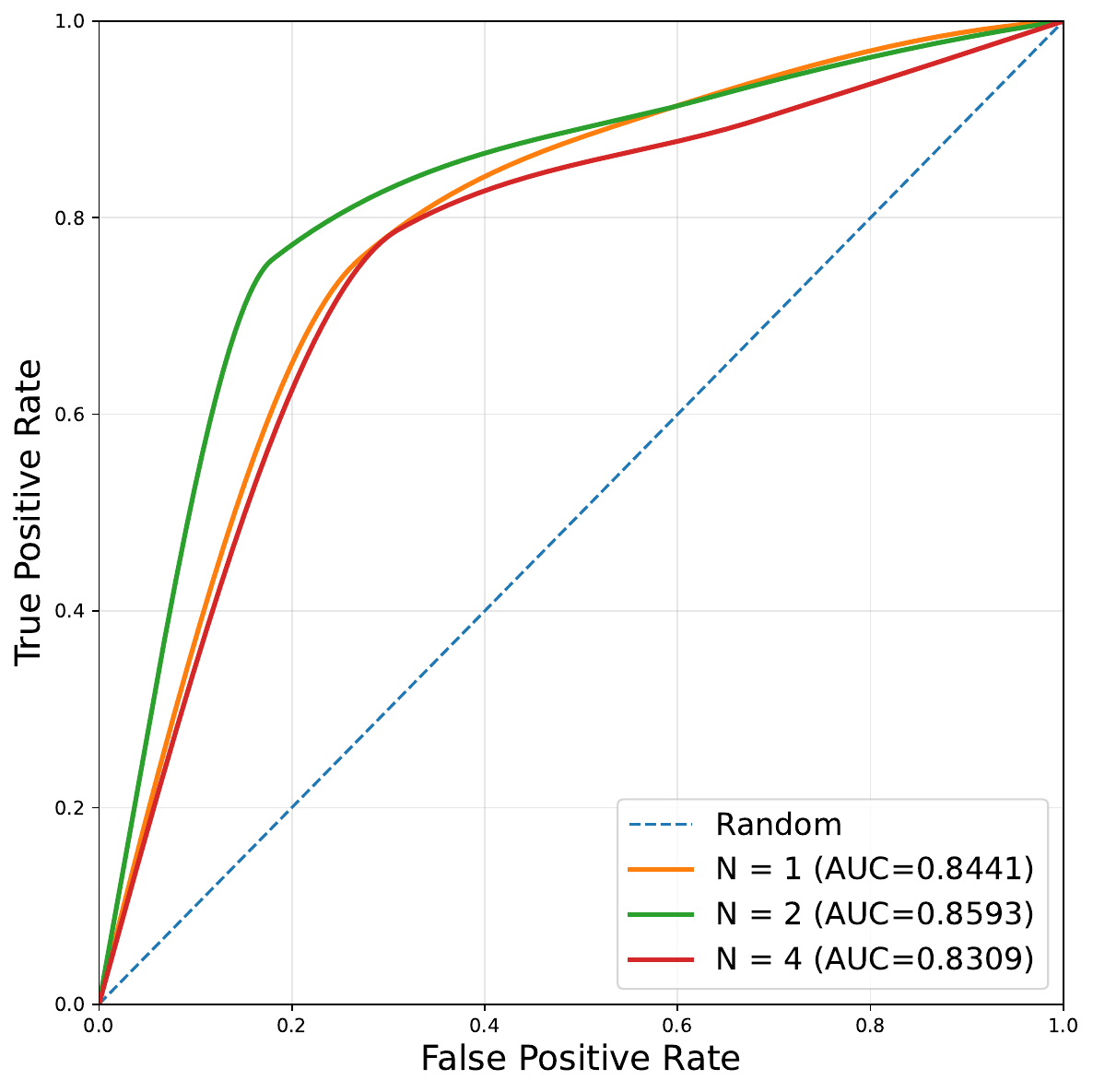}
    \caption{}\label{fig:fusion_blocks_auc}
\end{subfigure}

\caption{(a) Effect of prompt depth on F1-score for the VinDr-Mammo 5-class classification task. (b) AUC-ROC performance on the VinDr-Mammo binary classification task for different numbers of cross-view fusion blocks, where $N$ denotes the number of fusion blocks.}
\label{fig:ablation}
\end{figure}

\textbf{Prompt Depth.} We evaluate the effect of inserting learnable prompts at different depths of the MedSigLIP transformer. As shown in Figure~\ref{fig:prompt_depth}, prompt depth has a substantial impact on classification performance. The best result is obtained at depth 12, achieving an F1-score of 47.66\% on the VinDr-Mammo 5-class task. Comparable performance is observed at depths 8 and 16, while both shallower and deeper prompt placements lead to noticeable degradation. In particular, performance drops considerably at depths 20, 24, and 27, indicating that extending prompt-based adaptation to a larger number of transformer layers is not consistently beneficial. This finding motivates the integration of explicit cross-view fusion alongside prompt-based adaptation.

\textbf{Number of Fusion Blocks.} We investigate the effect of the number and placement of fusion blocks in the proposed framework. As shown in Figure~\ref{fig:fusion_blocks_auc}, experiments are conducted on the VinDr-Mammo binary classification task using different fusion configurations over MedSigLIP transformer layers, where layer indexing starts from 0. In this setting, the prompt depth is fixed to 8, and fusion blocks are inserted at selected layers to enable cross-view interaction at different stages of the encoder.

We first evaluate a single fusion block placed at layer 18, which achieves an AUC of 0.8441. Next, we introduce two fusion blocks placed at layers 12 and 23, resulting in the best performance of 0.8593. Finally, we evaluate a configuration with four fusion blocks at layers 12, 18, 23, and 26 (final layer), which does not improve performance and yields an AUC of 0.8313, suggesting that overly dense fusion across deeper layers may not be beneficial. Overall, the results indicate that a moderate number of well-spaced fusion blocks provides the most effective balance for cross-view feature integration.

\textbf{Summary.}
The ablation results show that each component contributes to the overall performance. The fusion-token mechanism consistently improves results over residual fusion, while prompt configuration and encoder fine-tuning further influence performance. These results support the effectiveness of the proposed design choices under the evaluated settings.

\begin{figure}[pos=htbp]
\centering
\includegraphics[width=0.95\linewidth]{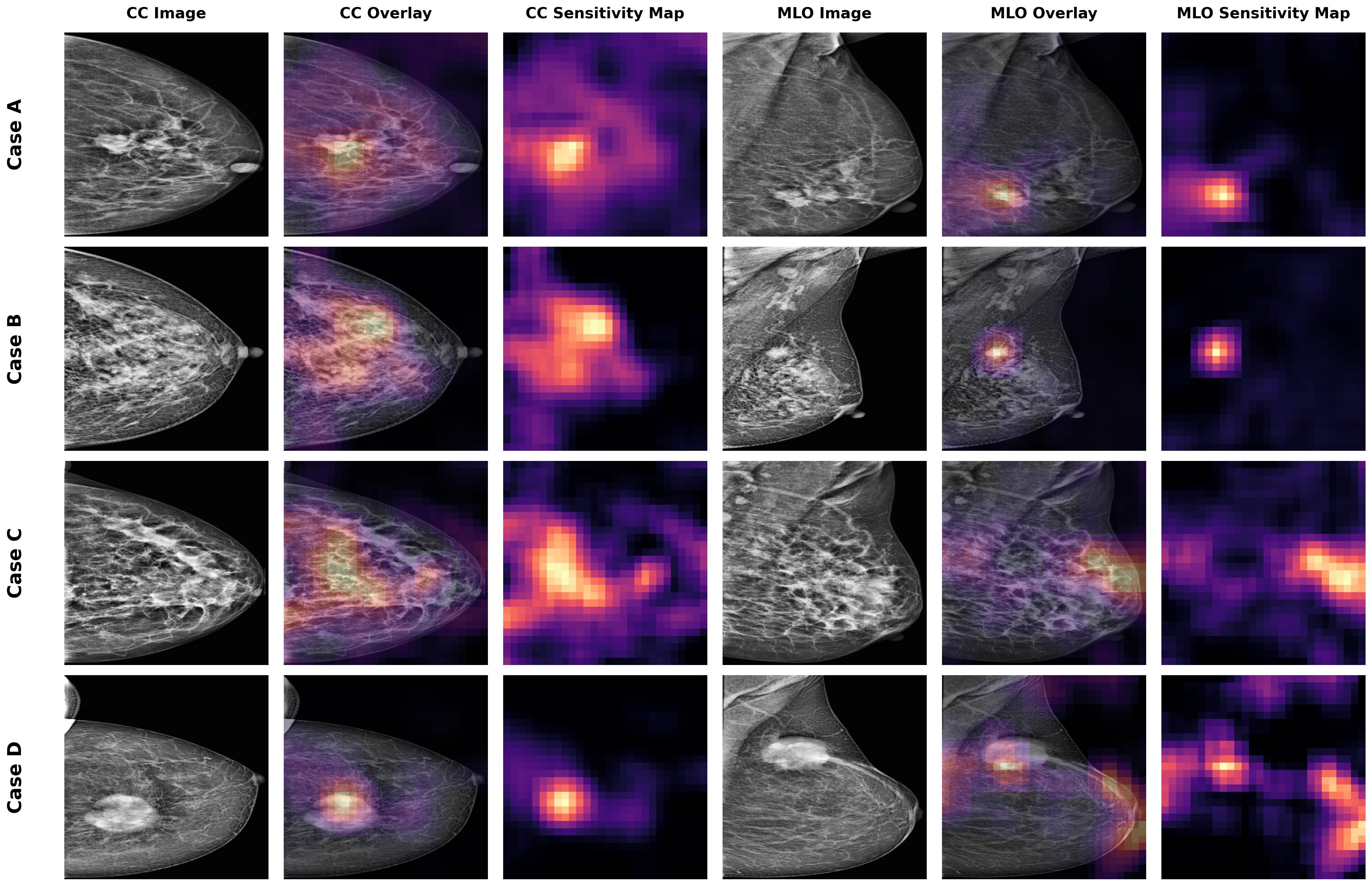}
\caption{Qualitative interpretability results obtained using occlusion-based sensitivity analysis~\cite{zeiler2014visualizing}. From left to right, the columns present the original craniocaudal (CC) image, the CC overlay, the CC sensitivity map, the original mediolateral oblique (MLO) image, the MLO overlay, and the MLO sensitivity map. Regions with warmer colors correspond to image areas whose occlusion leads to a larger reduction in the predicted class score, indicating stronger influence on the model's decision.}
\label{fig:vis}
\end{figure}

Figure \ref{fig:vis} illustrates the model’s interpretability using occlusion-based sensitivity analysis \cite{zeiler2014visualizing}, highlighting the regions most influential to the prediction.

\section{Conclusion}

This work investigated how large pretrained vision transformers can be effectively adapted for multi-view mammography classification while maintaining a frozen backbone. We showed that adaptation and cross-view integration can be unified within a token-centric framework, where prompt tokens and cross-view fusion tokens introduce task-specific and view-specific information into pretrained representations. By combining shared prompt adaptation with progressive cross-view interaction at intermediate transformer depths, the proposed framework enables effective integration of complementary information from CC and MLO views throughout the representation hierarchy. Experimental results on the VinDr-Mammo and CMMD datasets demonstrate consistent improvements over linear probing, prompt-only adaptation, and conventional fusion baselines. 

Beyond mammography, these findings highlight token-based adaptation and fusion as a flexible and scalable paradigm for leveraging vision foundation models in multi-view medical imaging, offering a promising alternative to conventional feature-level fusion strategies.

\section*{Code Availability}
The source code is publicly available at:
https://github.com/PartAI-Projects/CrossViewTokenFusion
\section*{Acknowledgment}

The authors acknowledge the financial support of the PART AI Research Center, which contributed to the successful completion of this research.

\bibliographystyle{cas-model2-names}

\bibliography{cas-refs}

\end{document}